\pdfoutput=1

\documentclass[11pt]{article}

\usepackage{EACL2023}

\usepackage{times}
\usepackage{latexsym}

\usepackage[T1]{fontenc}

\usepackage[utf8]{inputenc}

\usepackage{microtype}

\usepackage{inconsolata}

\usepackage[dvipsnames]{xcolor}
\usepackage{multirow}
\usepackage{booktabs}
\usepackage{xspace}
\usepackage{graphicx}
\usepackage{subcaption}
\usepackage{listings}

\usepackage{xcolor,colortbl}
\definecolor{Gray}{gray}{0.85}
\newcolumntype{g}{>{\columncolor{Gray}}c}

\newcommand{\numLanguages}{17\xspace}

\newcommand{\heideltime}{HeidelTime\xspace}
\newcommand{\spacy}{spacy\xspace}
\newcommand{\systemExtMono}{Mono}
\newcommand{\systemExtMulti}{Multi}
\newcommand{\systemNorm}{\textsc{Our}}

\newcommand{\systemMono}{\systemExtMono+\systemNorm}
\newcommand{\systemMulti}{\systemExtMulti+\systemNorm}
\newcommand{\systemGold}{Gold+\systemNorm}
\newcommand{\fscore}[1][1]{$F_{#1}$\xspace}

\title{Multilingual Normalization of Temporal Expressions \\with Masked Language Models}

\author{Lukas Lange$^{1}$ \\
	\And
	Jannik Str\"{o}tgen$^1$ \\
	\hspace{4cm}$^1$ Bosch Center for Artificial Intelligence, Renningen, Germany\\
	\hspace{4cm}$^2$ Spoken Language Systems (LSV), Saarland University, Saarbr\"{u}cken, Germany\\
	{\tt \hspace{4cm}\{Lukas.Lange,Jannik.Stroetgen,Heike.Adel\}@de.bosch.com} \\
	{\tt \hspace{4cm}dietrich.klakow@lsv.uni-saarland.de} \\
	\And
	Heike Adel$^1$ \\
	\And
	Dietrich Klakow$^2$ \\
	\\}

\begin{document}
\maketitle
\begin{abstract}
The detection and normalization of temporal expressions is an important task and preprocessing step for many applications. 
However, prior work on normalization is rule-based, which severely limits the applicability in real-world multilingual settings, due to the costly creation of new rules. 
We propose a novel neural method for normalizing temporal expressions based on masked language modeling. 
Our multilingual method outperforms prior rule-based systems in many languages, and in particular, for low-resource languages with performance improvements of up to 33 \fscore on average compared to the state of the art.  
\end{abstract}

\section{Introduction}
Temporal tagging consists of the extraction of temporal expressions (TE) from texts and their 
normalization to a standard format (e.g., \emph{May '22}: \texttt{2022-05}).
While there are deep-learning approaches for the extraction, temporal tagging as a whole is usually solved with highly specific rule-based systems, 
such as SUTime \cite{Chang:Manning:LREC:2012:Temporal:SUTime} or HeidelTime \cite{Strotgen:Gertz:LRE:2013:Temporal:MultilingDomainTagging}. 
However, transferring rule-based methods to new languages or text domains 
requires a large manual effort to create 
rules specific to the target language.
Although work on the automatic generation of 
rules 
for many languages \cite{Streotgen:Gertz:EMNLP:2015:Temporal:BaselineTemporalTagger} exists, 
the rule quality 
typically does not match the high accuracy of 
hand-crafted rules. 

In contrast to rule-based systems, neural networks are known for their ability to generalize to new targets, in particular, for cross- and multilingual applications \cite{Rahimi:Li:ACL:2019:NER:MassivelyMultilingTransfer,Artexte:Schwenk:TACL:2021:Embeddings:LASER}. In the context of temporal tagging, recent works have shown 
promising results of neural networks for TE extraction 
in monolingual \cite{Laparra:Xu:TACL:2018:Temporal:NeuralTempNER} and multilingual settings \cite{Lange:Iurshina:RepL:2020:Adversarial/Temporal:MultilingualExtraction} where a single neural model is trained on many languages at once. 
However, TE normalization remains challenging, and no 
solution for the normalization 
across languages 
exists yet.

\begin{figure}
    \centering
    \includegraphics[width=.5\textwidth]{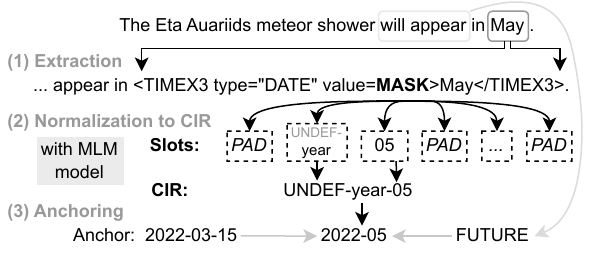}
    \caption{Overview of our 3-step pipeline for temporal tagging consisting of extraction of temporal expressions (step 1), normalization to a context-independent representation (CIR) using a slot-based masked language model (step 2) and anchoring given a reference time and further contextual cues (step 3).}
    \label{fig:example}
\end{figure}

We propose a new multilingual normalization method which can make use of labeled data from many languages by training a neural transformer model with a masked language modeling (MLM) objective. 
Thus, we adopt the MLM objective function for a new purpose: TE normalization.

To the best of our knowledge, this is the first work that uses neural networks for TE normalization. 
For this, as shown in Figure~\ref{fig:example} and detailed below, we split the normalization task into two steps: normalization to a context-independent representation (CIR) and anchoring this representation using the document context.

The main contribution of this paper is our novel neural normalization method based on masked language modeling. 
For this, we create a large-scale multilingual dataset with weakly-supervised annotations of TEs and their normalized values in 87 languages. 
Our extensive set of experiments across \numLanguages languages demonstrates that our multilingual method robustly works for many languages and outperforms the state of the art for multilingual temporal tagging, \heideltime~\cite{Streotgen:Gertz:EMNLP:2015:Temporal:BaselineTemporalTagger}, 
especially for low-resource languages by more than 33 \fscore on average.
Further, we explore different training and decoding strategies for our model. 
The code for our models and the weakly- supervised data is publicly available.\footnote{\url{https://github.com/boschresearch/temporal-tagging-eacl}}

\section{Related Work}
\textbf{TE Normalization. }Besides rule-based systems \cite{Chang:Manning:LREC:2012:Temporal:SUTime,Strotgen:Gertz:LRE:2013:Temporal:MultilingDomainTagging}, 
one normalization method for TEs are context-free grammars \cite{Bethard:EMNLP:2013:Temporal:GrammarNormalization,Lee:Artzi:ACL:2014:Temporal:UWTime} which are independent of the extraction method. However, they are even more language-specific than rule-based systems and hardly generalizable to new languages. 
\citet{Laparra:Xu:TACL:2018:Temporal:NeuralTempNER} used a rule-based procedure for English TE normalization based on the SCATE format proposed by \citet{Bethard:Parker:LREC:2016:Temporal:SCATE}.
While their method could be extended to multilingual applications, no annotated data for other languages is available in the SCATE format, and it is mostly incompatible with the predominant TimeML \cite{Pustejovsky:Ingria:OUP:2005:Temporal:TimeML} annotation format. Therefore, we will focus on the TimeML format in this work and present the first neural approach to TE normalization.

\textbf{Masked Language Modeling (MLM).}
The MLM paradigm gains a lot of attention \cite{Sun:Liu:Arxiv:2021:General:Paradigmshift} due to popular language models like BERT~\cite{Devlin:Chang:NAACL:2019:Embeddings:BERT}
This leads to active research on using MLM to solve further tasks like 
text classification~\cite{Brown:Mann:NIPS:2020:Embeddings:GPT3}, named entity recognition~\cite{Ma:Zhou:Arxiv:2021:NER:Prompt} and relation extraction~\cite{Han:Zhao:Arxiv:General:PromptTextClass}, also in low-resource languages~\citep{hedderich-etal-2021-survey}. 
In this work, we adopt it to TE normalization for the first time.

\section{Background on Temporal Tagging}\label{sec:background}
Temporal tagging addresses the detection, classification and normalization of temporal expressions in unstructured texts --- often following the 
TimeML specifications~\cite{Pustejovsky:Ingria:OUP:2005:Temporal:TimeML}.

TimeML's most important attributes are \texttt{type} (the class of an expression, e.g., \textsc{Date}, \textsc{Time}, \textsc{Duration} or \textsc{Set}), and \texttt{value} (the normalized meaning of an expression, e.g., \texttt{YYYY-MM-DD} for specific days, such as \texttt{2022-05-01} for May 1, 2022).
While some TEs contain all necessary information for the normalization, e.g., ``May 1, 2022'', many expressions are incomplete w.r.t. the temporal information required for a normalization. An example is a \textit{relative expression} like ``yesterday'' which needs an anchor point. Given the anchor point May 1, 2022, for example from the document creation time, ``yesterday'' should be annotated with \texttt{type}=\textsc{Date} and \texttt{value}=\texttt{2022-04-30}.

Determining the anchor point can be challenging as it requires additional 
context information that could be given anywhere in the document. 
Therefore, systems for TE normalization, such as \heideltime~\cite{Strotgen:Gertz:LRE:2013:Temporal:MultilingDomainTagging}, create an intermediate context-independent representation (CIR) of the \texttt{value}. In the syntax of \heideltime, the expression \emph{yesterday} would result in a CIR of \texttt{UNDEF-last-day}. Similarly, an \textit{underspecified expression}, such as ``May'' would be represented with a CIR of \texttt{UNDEF-year-05}. Note that such a syntax for CIRs is language-independent. See Appendix~\ref{sec:appendix:values} for more details. To determine the final \texttt{value}, the CIR needs to be anchored given, e.g., a reference date and further cues (such as tense information).

\section{Approach}
\label{sec:approach}
We propose to approach multilingual temporal tagging in three 
steps as shown in Figure~\ref{fig:example}:
(1) Extraction of temporal expressions and their types using a multilingual sequence tagger; 
(2) Normalization of TEs to CIRs with our novel MLM-based normalization model; 
(3) Anchoring of CIRs given a reference time, e.g., using \heideltime rules.

Our main contribution is a neural model for the second subtask, the normalization to a CIR. 
To the best of our knowledge, this has not been addressed with neural networks before.
In this section, we detail all components of our approach. Information on the models that we apply for the first and third subtasks as well as an ablation study of directly predicting the normalized anchored value (without CIRs) are given in Section \ref{sec:experiments}. 

\paragraph{Masked Language Modeling. }
We model the task of assigning CIRs to temporal expressions as masked language modeling. In particular, we add TimeML annotations as inline information to the text sequences and mask the value field for prediction, e.g.,
"... <TIMEX3 type="DATE" value="\textbf{MASK}">yesterday</TIMEX3> ...". Note that those annotations could be the ground-truth annotations when applying the model on gold temporal expressions or predicted temporal expressions when using the model in the 3-step pipeline as described above.
In our experiments, we train a transformer model for CIR prediction using the masked language modeling (MLM) objective.

\paragraph{Slot-Based Value Representation. }
Using only a single mask token for the whole CIR would require the model to store all possible CIRs in its vocabulary. Since it is not possible to enumerate, i.a., all possible dates, we model the CIRs as a fixed-length sequence of slots. 
In particular, we define 11 slots and use regular expressions to split the value field into slots in the training data. 
Figure \ref{fig:example} shows an example for the CIR ``UNDEF-year-05'' that is represented as the slots `` [PAD], [year], [05], [PAD], ..., [PAD]''. Details on the slots and regular expressions are given in Appendix~\ref{sec:appendix:values}.
To cover the full vocabulary of CIRs, we introduce 200 new tokens to the vocabulary of the 
language model.\footnote{In our experiments, we compare 
our pre-defined slots to using subtokens from the language model tokenizer.}

\paragraph{Curriculum Learning. }
Our slot-based representation with 11 slots per CIR results in 11 masks.
To train the model on this task, we apply curriculum learning in the first half of the training. In particular, we start with masking only a single slot of the CIR and steadily increase the number of masks up to the maximum of 11. For the second half of the training, masking is applied to all slots. We follow \citet{Devlin:Chang:NAACL:2019:Embeddings:BERT} and mask different parts of the input with different probabilities. In particular, we mask the value slots with a probability of 70\%, annotated tokens with 15\%, types with 10\% and other text parts with 5\%.

\paragraph{Inference and Decoding. }
For inference, we first add 11 masks (i.e., one per slot) to the input sentence. They serve as value placeholders that need to be predicted. Then, we use the masked language model to predict the most probable sequence of slots for the CIR. 
To decode the sequence, we apply sequential left-to-right decoding of all masks by iteratively decoding the left-most mask and replacing the mask with its predicted value until all masks are resolved.
We compare this to two alternative decoding strategies: (i) decoding all masks simultaneously, (ii) training a conditional random field model that takes the logits as input and uses the Viterbi algorithm to determine the most probable sequence of predictions~\cite{Lafferty:MaCallumg:ICML:2001:General:CRF}. 


\section{Experiments}
\label{sec:experiments}
This section describes our experiments and discusses the results. We compare our model to \heideltime~\cite{Strotgen:Gertz:LRE:2013:Temporal:MultilingDomainTagging}, the current state of the art for multilingual temporal tagging. For evaluation, we use the TempEval3 evaluation script \cite{Uzzaman:Llorens:SEM:2013:Temporal:TempEval3} and report strict, relaxed and type \fscore for the extraction and value \fscore for our normalization experiments, respectively.

\paragraph{Evaluation Data. }
Our models are evaluated on gold-standard corpora in \numLanguages languages.
Details on the corpora are given in Appendix~\ref{app:sec:gold}. We divide the languages into high- and low-resource depending on whether manually created \heideltime rules are available for the respective language.  

\paragraph{Training Data. }
For training the normalization model, we create a large-scale weakly-supervised dataset covering 87 languages.\footnote{The set of 87 languages is the intersection of languages covered by \heideltime, our data and the XLM-R language model that we use for initializing our models. }
Reasons are that (i) existing gold training data is too small to cover the wide range of different values and 
(ii) CIRs are not part of existing annotations.
For all languages, we take the data from GlobalVoices\footnote{\url{https://globalvoices.org/}} (news-style documents) and Wikipedia\footnote{\url{https://wikipedia.org/}} 
(narrative-style documents), use \spacy
for tokenization and \heideltime for the annotation with temporal expressions. 
The number and quality of annotations is highly dependent on the amount of available data for that language and the quality of \heideltime's rules.
Details on the weakly-supervised data are given in Appendix~\ref{app:sec:weak}.

%

\begin{table*}
\centering
\footnotesize
\begin{tabular}{l cccg cccg cccg c} \toprule
\multirow{2}{*}{} & \multicolumn{4}{c}{\bf \heideltime} & \multicolumn{4}{c}{\bf \systemMono} & \multicolumn{4}{c}{\bf \systemMulti} & \multicolumn{1}{c}{\bf \systemGold} \\
  & \textbf{Str.} & \textbf{Rel.} & \textbf{Type} & \textbf{Val.} & \textbf{Str.} & \textbf{Rel.} & \textbf{Type} & \textbf{Val.} & \textbf{Str.} & \textbf{Rel.} & \textbf{Type} & \textbf{Val.} & \textbf{Val.} \\
  \cmidrule(lr){2-5} \cmidrule(lr){6-9} \cmidrule(lr){10-13} \cmidrule(lr){14-14} 
de (N) & 69.7 & 79.3 & 75.4 & \textbf{62.4} & \textbf{75.4} & \textbf{85.9} & \textbf{80.6} & 61.5 & 70.9 & 82.6 & 76.2 & 59.5 & 74.1 \\
de (W) & 88.5 & 94.3 & 89.0 & 84.8 & \textbf{89.6} & \textbf{97.0} & \textbf{96.0} & 83.8 & 88.9 & 96.7 & 95.4 & \textbf{85.7} & 87.5 \\
en (N) & 81.8 & 90.7 & 83.3 & \textbf{78.1} & \textbf{85.7} & \textbf{92.3} & \textbf{86.5} & 72.5 & 82.0 & 88.9 & 82.8 & 70.5 & 79.0 \\
en (W) & 90.6 & 94.3 & 90.6 & \textbf{94.3} & 93.1 & 96.6 & \textbf{93.1} & 89.7 & \textbf{94.7} & \textbf{98.3} & 87.7 & 94.2 & 94.2 \\
es (N) & 83.7 & 90.2 & 86.1 & \textbf{80.9} & \textbf{89.6} & \textbf{94.5} & \textbf{91.4} & 79.0 & 89.3 & 94.2 & 90.0 & 77.1 & 84.4 \\
et (N) & 42.4 & 57.4 & 51.3 & 44.0 &  3.3 & 28.0 & 24.4 &  9.6 & \textbf{55.5} & \textbf{78.0} & \textbf{72.0} & \textbf{45.2} & 64.8 \\
fr (N) & \textbf{85.6} & \textbf{90.6} & \textbf{82.3} & \textbf{73.3} & 82.5 & 88.1 & 79.7 & 67.9 & 82.4 & 89.8 & 76.9 & 61.4 & 68.0 \\
hr (W) & \textbf{93.3} & \textbf{95.8} & \textbf{94.6} & \textbf{85.7} & 84.1 & 90.8 & 89.5 & 74.6 & 86.3 & 91.7 & 90.1 & 75.7 & 84.7 \\
it (N) & \textbf{84.4} & \textbf{92.9} & \textbf{83.5} & \textbf{74.1} & 69.8 & 81.4 & 73.7 & 60.4 & 76.8 & 82.4 & 78.4 & 67.2 & 75.3 \\
nl (N) & 54.0 & \textbf{91.3} & 79.0 & 44.4 & 61.4 & 73.0 & 67.2 & 42.7 &\textbf{76.0} & 82.7 & \textbf{81.4} & \textbf{53.5} & 64.6  \\
pt (N) & 71.3 & 80.9 & 76.5 & 63.2 & \textbf{87.1} & \textbf{91.2} & 85.0 & \textbf{68.7} & \textbf{87.1} & 91.1 & \textbf{86.5} & \textbf{68.7} & 76.6 \\
vi (W) & \textbf{92.6} & 89.5 & \textbf{96.6} & \textbf{91.6} & 87.6 & 85.0 & 89.8 & 83.5 & 91.5 & \textbf{93.8} & 92.6 & 90.8 & 91.2 \\ 
\cmidrule(lr){2-14}
\textbf{avg.} & \textbf{78.2} & 87.3 & 82.4 & \textbf{75.6} & 75.8 & 83.7 & 79.7 & 66.2 & 81.8 & \textbf{89.2} & \textbf{84.2} & 70.7 & 78.7 \\
\midrule \midrule
\multirow{2}{*}{} & \multicolumn{4}{c}{\bf \heideltime-auto} & \multicolumn{4}{c}{\bf \systemMono} & \multicolumn{4}{c}{\bf \systemMulti} & \multicolumn{1}{c}{\bf \systemGold} \\
  & \textbf{Str.} & \textbf{Rel.} & \textbf{Type} & \textbf{Val.} & \textbf{Str.} & \textbf{Rel.} & \textbf{Type} & \textbf{Val.} & \textbf{Str.} & \textbf{Rel.} & \textbf{Type} & \textbf{Val.} & \textbf{Val.} \\
  \cmidrule(lr){2-5} \cmidrule(lr){6-9} \cmidrule(lr){10-13} \cmidrule(lr){14-14} 
ca (N) & 28.1 & 62.8 & 61.1 & 43.6 & 29.5 & 64.3 & 62.3 & 40.2 & \textbf{77.3} & \textbf{87.8} & \textbf{82.5} & \textbf{59.7} & 67.9 \\
el (W) & 2.2 & 4.9 & 4.9 & 1.3 & 47.0 & 88.2 & 86.1 & 64.6 & \textbf{81.7} & \textbf{92.0} & \textbf{90.2} & \textbf{70.6} & 83.7 \\
eu (N) & 22.5 & 26.8 & 23.9 & 18.3 &  0.0 &  0.0 &  0.0 &  0.0 & \textbf{59.7} & \textbf{70.2} & \textbf{66.0} & \textbf{45.0} & 51.2 \\
id (N) & 19.7 & 54.7 & 44.5 & 40.1 & 17.4 & 39.7 & 30.6 & 25.6 & \textbf{49.7} & \textbf{79.5} & \textbf{63.9} & \textbf{46.9} & 64.8 \\
pl (N) & 18.8 & 27.2 & 16.5 & 11.2 & 86.1 & \textbf{92.5} & 87.6 & 58.7 & \textbf{86.7} & 92.2 & \textbf{87.7} & \textbf{59.0} & 66.0 \\
ro (N) &  3.2 & 19.5 & 16.7 &  5.5 &  3.8 & 22.6 & 37.0 &  7.7 &  \textbf{9.8} & \textbf{47.2} & \textbf{39.1} & \textbf{19.7} & 54.6 \\
ua (W) &  1.6 &  2.8 &  2.2 &  1.2 & \textbf{80.2} & 90.6 & 87.5 & 63.6 & 79.4 & \textbf{90.7} & \textbf{88.8} & \textbf{65.4} & 74.5 \\
\cmidrule(lr){2-14}
\textbf{avg.} & 12.7 & 28.4 & 24.3 & 17.3 & 37.7 & 56.8 & 55.9 & 37.2 & \textbf{63.5} & \textbf{79.9} & \textbf{74.0} & \textbf{50.9} & 66.1 \\ 
\bottomrule
\end{tabular}
\caption{Detailed overview of our results for extraction (Str., Rel., Typ.) and normalization (Val.) per language for \textbf{N}ews and \textbf{W}iki domains. The upper and lower parts display high- and low-resource languages, respectively. 
}
\label{tab:results_detail}
\vspace{-0.13cm}
\end{table*}

\paragraph{3-Step Pipeline for Temporal Tagging. }
Both our temporal expression extraction and normalization models are based on the mulitlingual XLM-R transformer~\cite{Conneau:Khandelwal:ACL:2020:Embeddings:XLMR}.\footnote{\texttt{xlm-roberta-base} with 270M parameters.}

We model the \emph{TE 
extraction} as a sequence-labeling problem following \citet{Lange:Iurshina:RepL:2020:Adversarial/Temporal:MultilingualExtraction}. For this, we convert the annotated corpora 
into the BIO format. For the monolingual setting (\systemExtMono), we train one model per language on the 
gold-standard resources if available or the weakly-supervised data otherwise. For the multilingual setting (\systemExtMulti), we train a single model on the combined training resources of all languages.

For the \emph{normalization to CIRs}, we train our proposed model 
with masked language modeling (see Section \ref{sec:approach}).
In our experiments, we evaluate this model in combination with the multilingual extraction model (\systemMulti) as well as in combination with the gold boundaries for temporal expressions (\systemGold) which 
serves as an upper bound.

For \emph{anchoring CIRs}, we use rules similar to \heideltime's rules.\footnote{More precisely, we use a slightly modified version of \heideltime's \textsc{specifyAmbiguousValuesString} function which incorporates tense information of the context using morphological features from \spacy (\url{https://spacy.io/usage/linguistic-features\#morphology}).}
In particular, anchor dates can be given by the document creation time or by previous temporal expressions~\cite{Strotgen:Gertz:Book:2016:Temporal:Tagging}.

\paragraph{Results. } 
Table~\ref{tab:results_detail} gives an overview of our experimental results.
Multilingual extraction outperforms monolingual extraction, probably because the model is able to use knowledge from different languages. 
Our multilingual model achieves +2 \fscore for high-resource and +51 \fscore for low-resource languages compared to \heideltime. 

The normalization results are given in the \textit{Val.} columns of Table~\ref{tab:results_detail}. 
Our masked language model is matching \heideltime's performance rather close for high-resource languages and outperforms it for low-resource languages with an increase of 33 \fscore points on average with our multilingual extraction model. Note that our models are multilingual, thus, we can use the same model for all languages.\footnote{Since we actually train the MLM model on 87 languages, we could even apply it to more languages if there were gold-standard evaluation datasets publicly available.}
The upper bound of using gold extractions (\systemGold) shows that the extraction part still offers 
room for future improvements. 

Note that \heideltime with automatically created rules has a poor performance for some low-resource languages (el, ro, ua). This is similar to the observations by \citet{Grabar:Hamon:Article:2019:Temporal:WikiwarsUA} who found that ``[e]xploitation of this automatically built system produced no results when applied to the Ukrainian data.''
For those languages, the automatic rule generation is not good enough in practice
which emphasizes the need for multilingual systems like our model.

\paragraph{Ablation Studies.}\label{sec:ablation}
As our proposed model consists of multiple components, we now investigate their individual effects in more detail. The results for our ablation studies are given in Table~\ref{tab:ablation}.  

First, we test different \emph{decoding strategies} as described in Section \ref{sec:approach}.
We find that sequential decoding works best. However, it also requires more computation time. A cheaper alternative with only minor performance decreases is the simultaneous decoding of all masks. 

Second, we analyze the impact of different \emph{value representations} by comparing our proposed approach with CIR and slot tokenization to (i) tokenization of values using the standard XLM-R tokenizer instead of pre-defined slots (w/o \textsc{Our} Slots), and (ii) training a model to directly predict the anchored value without CIRs in between (w/o \textsc{Our} CIR).
For (i), we find that our slot method has major advantages when processing narrative texts, such as Wikipedia, due to the higher amount of relative expressions (cf., Table~\ref{tab:value_distribution} in Appendix~\ref{sec:app:dist}), that are tokenized into many subtokens (up to 34, instead of 11 when using our slots). 
For (ii), we add the document creation time to the input 
so that the model has more temporal information to predict the fully normalized value directly instead of a CIR. 
However, we find that current transformers are not able to correctly incorporate this information in a combined normalizing+anchoring step and mostly predict a memorized, incorrect value. Thus, using CIRs as an intermediate step is important for neural temporal tagging.

Third, we investigate the \emph{training strategy} and \emph{training data}. Our curriculum learning has advantages for low-resource languages as it reduces the training complexity which helps for the difficult adaptation to languages with few resources. 
Weakly-supervised training data is required, as the amount of gold-standard data is too small to train the MLM model. Finetuning the trained MLM model further on gold data (Weak+Gold) decreases performance slightly. 
Training the model on monolingual data only also decreases performance, highlighting the prospects of our multilingual approach. 

Finally, we compare our models to an \emph{encoder-decoder model}, i.e., an autoregressive language model that we adapt to TE normalization. 
For this, we follow the entity linking approach from \citet{decao2021autoregressive} and train a BART encoder-decoder model \citep{lewis-etal-2020-bart} for constrained decoding against a subspace of normalized TEs with our weakly-supervised data. We add the document creation time to the input, mark the extracted annotations and keep other TEs in the context, as in our other experiments.
Given the gigantic amount of possible temporal expressions, e.g., there are roughly 32M seconds in a single year,
we have to prune the search space to a reasonable size. 
Thus, we do not use time expressions of hour and smaller granularities and restrict the search space to years and months from 1 AD to 2100. Finer elements like weeks, days and daytimes are added for years between 2000 and 2026. We use durations for all defined units with numbers up to 10,000, e.g., 10,000 days.
With this, we prune the search space to 1.4B terms which we store in a prefix tree. This results in an acceptable inference speed with BEAM search (5 beams). It takes roughly twice as long as our sequential MLM decoding. Note that this BART model has more parameters (400M) than the base version of XLM-R (270M) that we use in our model. 
The results are shown in the lower part of Table~\ref{tab:ablation}. We see, that our proposed MLM normalization model outperforms the BART model by a large margin. Nonetheless, the encoder-decoder model performs comparable to our model variant that directly predicts fully-normalized expressions. This clearly highlights the need for normalizing to CIRs before anchoring temporal expressions.

\begin{table}
\setlength\tabcolsep{4.2pt}
\centering
\footnotesize
\begin{tabular}{l cc cc cc} \toprule
 & \multicolumn{2}{c}{\bf News} & \multicolumn{2}{c}{\bf Wiki} & \multicolumn{2}{c}{\bf Low-R.} \\
 & \textbf{de} & \textbf{en} & \textbf{de} & \textbf{en} & \textbf{ca} & \textbf{eu} \\
  \cmidrule(lr){2-3} \cmidrule(lr){4-5} \cmidrule(lr){6-7}
\systemNorm & \textbf{74.1} & 79.0 & \textbf{87.5} & 94.2 & 67.9 & \textbf{51.2} \\ 
  \cmidrule(lr){2-3} \cmidrule(lr){4-5} \cmidrule(lr){6-7}
\multicolumn{7}{l}{\quad \textit{Decoding Strategy} (\textsc{Our} uses Sequential)} \\
w/ Simultaneous          & 73.3 & 78.3 & \textbf{87.5} & 93.5 & \textbf{68.1} & 50.4 \\
w/ Viterbi               & 73.3 & 78.3 & \textbf{87.5} & 94.2 & 67.9 & 50.4 \\ 
  \cmidrule(lr){2-3} \cmidrule(lr){4-5} \cmidrule(lr){6-7}
\multicolumn{7}{l}{\quad \textit{Value Representation}} \\
w/o \textsc{Our} Slots   & 71.9 & 77.9 & 83.3 & 92.8 & 63.7 & 27.8 \\
w/o \textsc{Our} CIR     & 68.5 & 68.0 & 66.5 & 55.7 & 41.6 & 21.2 \\ 
  \cmidrule(lr){2-3} \cmidrule(lr){4-5} \cmidrule(lr){6-7}
\multicolumn{7}{l}{\quad \textit{Training Strategy}} \\
w/o Curriculum           & 72.1 & \textbf{80.4} & 85.5 & \textbf{94.4} & 64.7 & 29.3 \\ 
  \cmidrule(lr){2-3} \cmidrule(lr){4-5} \cmidrule(lr){6-7}
\multicolumn{7}{l}{\quad \textit{Training Data} (\textsc{Our} uses Weak)} \\
Weak + Gold              & 68.3 & 76.1 & 58.2 & 93.3 & - & - \\
only Gold                & 14.2 & 13.8 &  6.7 &  3.6 & - & - \\ 
only Monolingual         & 63.1 & 76.8 & 86.8 & 91.4 & 29.2 & 8.9 \\ 
\midrule 
\multicolumn{7}{l}{\quad \textit{Encoder-Decoder Model}} \\
Monolingual  & 59.3 & 67.4 & 49.4 & 59.6 & 2.2 & 7.3 \\
Multilingual & 63.5 & 63.0 & 49.0 & 58.5 & 54.8 & 25.2 \\
\bottomrule
\end{tabular}
\caption{Ablation study for our model components (Value \fscore) on gold extractions. 
Low-R. stands for low-resource languages without gold training data. 
}
\label{tab:ablation}
\setlength\tabcolsep{6pt}
\end{table}

\section{Conclusion}
In this paper, we introduced a new method for normalizing temporal expressions based on masked language modeling and a new slot-based prediction scheme of context-independent representations. With this approach, we were able to train a single multilingual model for the task. 
We evaluated our method in \numLanguages languages and set the new state of the art in low-resource languages with massive improvements of 35 \fscore points on average. The success of our method demonstrates the potential of neural networks for temporal normalization and we are convinced that it will enable future research on this topic. An interesting research direction is the joint modeling of extraction and normalization. 

\section*{Limitations}
Our experiments are focused on Indo-European languages due to the lack of publicly available, labeled 
data points in other languages. Exceptions for which we could test zero-shot transfer were Basque, 
Estonian, Indonesian and Vietnamese. Even though, our model is working for these languages, it is not clear if the multilingual models transfer to all languages seen in the pre-training or by our weak supervision. 
The training of the multilingual models requires a considerable number of computational resources (up 
to 1.5 GPU days), which might not be available for all people/organizations. By publishing our model, we 
hope to lower the barrier for this kind of research by providing a pre-trained starting point. 
An in-depth error analysis to better understand which types of temporal expressions are well or less well 
covered in which language by our model was not performed. We are full of hope that such analyses can 
be tackled by users of our models who have the required language skills so that the analysis does not 
have to be limited to English.

\section*{Acknowledgments}
We would like to thank the members of the BCAI NLP \& NS-AI research group and the anonymous reviewers for their helpful comments.

\bibliography{custom}

\begin{thebibliography}{39}
\expandafter\ifx\csname natexlab\endcsname\relax\def\natexlab#1{#1}\fi

\bibitem[{Altuna et~al.(2020)Altuna, Aranzabe, and Díaz~de
  Ilarraza}]{Altuna:Aranzabe:Article:2020:TemporalEusTimeML}
Begoña Altuna, María~Jesús Aranzabe, and Arantza Díaz~de Ilarraza. 2020.
\newblock \href {https://doi.org/10.32714/ricl.08.01.06} {Eustimeml: A mark-up
  language for temporal information in basque}.
\newblock \emph{Research in Corpus Linguistics}, 8(1):86--104.

\bibitem[{Artetxe and
  Schwenk(2019)}]{Artexte:Schwenk:TACL:2021:Embeddings:LASER}
Mikel Artetxe and Holger Schwenk. 2019.
\newblock \href {https://doi.org/10.1162/tacl_a_00288} {{Massively Multilingual
  Sentence Embeddings for Zero-Shot Cross-Lingual Transfer and Beyond}}.
\newblock \emph{Transactions of the Association for Computational Linguistics},
  7:597--610.

\bibitem[{Bethard(2013)}]{Bethard:EMNLP:2013:Temporal:GrammarNormalization}
Steven Bethard. 2013.
\newblock \href {https://aclanthology.org/D13-1078} {A synchronous context free
  grammar for time normalization}.
\newblock In \emph{Proceedings of the 2013 Conference on Empirical Methods in
  Natural Language Processing}, pages 821--826, Seattle, Washington, USA.
  Association for Computational Linguistics.

\bibitem[{Bethard and Parker(2016)}]{Bethard:Parker:LREC:2016:Temporal:SCATE}
Steven Bethard and Jonathan Parker. 2016.
\newblock \href {https://aclanthology.org/L16-1599} {A semantically
  compositional annotation scheme for time normalization}.
\newblock In \emph{Proceedings of the Tenth International Conference on
  Language Resources and Evaluation ({LREC}'16)}, pages 3779--3786,
  Portoro{\v{z}}, Slovenia. European Language Resources Association (ELRA).

\bibitem[{Bittar et~al.(2011)Bittar, Amsili, Denis, and
  Danlos}]{Bittar:Amsili:ACL:2011:Temporal:FrenchTimeBank}
Andr{\'e} Bittar, Pascal Amsili, Pascal Denis, and Laurence Danlos. 2011.
\newblock \href {https://aclanthology.org/P11-2023} {{F}rench {T}ime{B}ank: An
  {ISO}-{T}ime{ML} annotated reference corpus}.
\newblock In \emph{Proceedings of the 49th Annual Meeting of the Association
  for Computational Linguistics: Human Language Technologies}, pages 130--134,
  Portland, Oregon, USA. Association for Computational Linguistics.

\bibitem[{Brown et~al.(2020)Brown, Mann, Ryder, Subbiah, Kaplan, Dhariwal,
  Neelakantan, Shyam, Sastry, Askell, Agarwal, Herbert-Voss, Krueger, Henighan,
  Child, Ramesh, Ziegler, Wu, Winter, Hesse, Chen, Sigler, Litwin, Gray, Chess,
  Clark, Berner, McCandlish, Radford, Sutskever, and
  Amodei}]{Brown:Mann:NIPS:2020:Embeddings:GPT3}
Tom Brown, Benjamin Mann, Nick Ryder, Melanie Subbiah, Jared~D Kaplan, Prafulla
  Dhariwal, Arvind Neelakantan, Pranav Shyam, Girish Sastry, Amanda Askell,
  Sandhini Agarwal, Ariel Herbert-Voss, Gretchen Krueger, Tom Henighan, Rewon
  Child, Aditya Ramesh, Daniel Ziegler, Jeffrey Wu, Clemens Winter, Chris
  Hesse, Mark Chen, Eric Sigler, Mateusz Litwin, Scott Gray, Benjamin Chess,
  Jack Clark, Christopher Berner, Sam McCandlish, Alec Radford, Ilya Sutskever,
  and Dario Amodei. 2020.
\newblock \href
  {https://proceedings.neurips.cc/paper/2020/file/1457c0d6bfcb4967418bfb8ac142f64a-Paper.pdf}
  {Language models are few-shot learners}.
\newblock In \emph{Advances in Neural Information Processing Systems},
  volume~33, pages 1877--1901. Curran Associates, Inc.

\bibitem[{Chang and Manning(2012)}]{Chang:Manning:LREC:2012:Temporal:SUTime}
Angel~X. Chang and Christopher Manning. 2012.
\newblock \href
  {http://www.lrec-conf.org/proceedings/lrec2012/pdf/284_Paper.pdf} {{SUT}ime:
  A library for recognizing and normalizing time expressions}.
\newblock In \emph{Proceedings of the Eighth International Conference on
  Language Resources and Evaluation ({LREC}'12)}.

\bibitem[{Conneau et~al.(2020)Conneau, Khandelwal, Goyal, Chaudhary, Wenzek,
  Guzm{\'a}n, Grave, Ott, Zettlemoyer, and
  Stoyanov}]{Conneau:Khandelwal:ACL:2020:Embeddings:XLMR}
Alexis Conneau, Kartikay Khandelwal, Naman Goyal, Vishrav Chaudhary, Guillaume
  Wenzek, Francisco Guzm{\'a}n, Edouard Grave, Myle Ott, Luke Zettlemoyer, and
  Veselin Stoyanov. 2020.
\newblock \href {https://www.aclweb.org/anthology/2020.acl-main.747}
  {Unsupervised cross-lingual representation learning at scale}.
\newblock In \emph{Proceedings of the 58th Annual Meeting of the Association
  for Computational Linguistics}, pages 8440--8451, Online. Association for
  Computational Linguistics.

\bibitem[{Costa and Branco(2012)}]{Costa:Branco:LREC:2012:Temporal:TimeBankPT}
Francisco Costa and Ant{\'o}nio Branco. 2012.
\newblock \href
  {http://www.lrec-conf.org/proceedings/lrec2012/pdf/246_Paper.pdf}
  {{T}ime{B}ank{PT}: A {T}ime{ML} annotated corpus of {P}ortuguese}.
\newblock In \emph{Proceedings of the Eighth International Conference on
  Language Resources and Evaluation ({LREC}'12)}, pages 3727--3734, Istanbul,
  Turkey. European Language Resources Association (ELRA).

\bibitem[{{De Cao} et~al.(2021){De Cao}, Izacard, Riedel, and
  Petroni}]{decao2021autoregressive}
Nicola {De Cao}, Gautier Izacard, Sebastian Riedel, and Fabio Petroni. 2021.
\newblock \href {https://openreview.net/forum?id=5k8F6UU39V} {Autoregressive
  entity retrieval}.
\newblock In \emph{9th International Conference on Learning Representations,
  {ICLR} 2021, Virtual Event, Austria, May 3-7, 2021}. OpenReview.net.

\bibitem[{Devlin et~al.(2019)Devlin, Chang, Lee, and
  Toutanova}]{Devlin:Chang:NAACL:2019:Embeddings:BERT}
Jacob Devlin, Ming-Wei Chang, Kenton Lee, and Kristina Toutanova. 2019.
\newblock \href {https://doi.org/10.18653/v1/N19-1423} {{BERT}: Pre-training of
  deep bidirectional transformers for language understanding}.
\newblock In \emph{Proceedings of the 2019 Conference of the North {A}merican
  Chapter of the Association for Computational Linguistics: Human Language
  Technologies, Volume 1 (Long and Short Papers)}, pages 4171--4186,
  Minneapolis, Minnesota. Association for Computational Linguistics.

\bibitem[{For{\u{a}}scu and
  Tufi{\c{s}}(2012)}]{Forascu:Tufis:LREC:2012:Temporal:RomanianTimeBank}
Corina For{\u{a}}scu and Dan Tufi{\c{s}}. 2012.
\newblock \href
  {http://www.lrec-conf.org/proceedings/lrec2012/pdf/770_Paper.pdf} {{R}omanian
  {T}ime{B}ank: An annotated parallel corpus for temporal information}.
\newblock In \emph{Proceedings of the Eighth International Conference on
  Language Resources and Evaluation ({LREC}'12)}, pages 3762--3766, Istanbul,
  Turkey. European Language Resources Association (ELRA).

\bibitem[{Grabar and
  Hamon(2019)}]{Grabar:Hamon:Article:2019:Temporal:WikiwarsUA}
Natalia Grabar and Thierry Hamon. 2019.
\newblock \href {http://ena.lp.edu.ua/handle/ntb/45492} {Wikiwars-ua: Ukrainian
  corpus annotated with temporal expressions}.
\newblock \emph{Computational Linguistics and Intelligent Systems}, 2:22--31.

\bibitem[{Han et~al.(2021)Han, Zhao, Ding, Liu, and
  Sun}]{Han:Zhao:Arxiv:General:PromptTextClass}
Xu~Han, Weilin Zhao, Ning Ding, Zhiyuan Liu, and Maosong Sun. 2021.
\newblock \href {https://arxiv.org/pdf/2105.11259.pdf} {Ptr: Prompt tuning with
  rules for text classification}.
\newblock \emph{arXiv preprint arXiv:2105.11259}.

\bibitem[{Hedderich et~al.(2021)Hedderich, Lange, Adel, Str{\"o}tgen, and
  Klakow}]{hedderich-etal-2021-survey}
Michael~A. Hedderich, Lukas Lange, Heike Adel, Jannik Str{\"o}tgen, and
  Dietrich Klakow. 2021.
\newblock \href {https://doi.org/10.18653/v1/2021.naacl-main.201} {A survey on
  recent approaches for natural language processing in low-resource scenarios}.
\newblock In \emph{Proceedings of the 2021 Conference of the North American
  Chapter of the Association for Computational Linguistics: Human Language
  Technologies}, pages 2545--2568, Online. Association for Computational
  Linguistics.

\bibitem[{Kapernaros(2020)}]{Kapernaros:Thesis:2020:Temporal:WikiWarsEL}
Emmanouil~I. Kapernaros. 2020.
\newblock \href
  {https://pergamos.lib.uoa.gr/uoa/dl/frontend/file/lib/default/data/2922239/theFile}
  {Extending the temporal tagger heideltime for the greek language}.

\bibitem[{Kocon et~al.(2019)Kocon, Oleksy, Bernas, and
  Marcinczuk}]{Kocon:Oleksy:Poleval:2019:Temporal:Overview}
Jan Kocon, Marcin Oleksy, Tomasz Bernas, and Micha{\l} Marcinczuk. 2019.
\newblock \href {http://poleval.pl/files/poleval2019.pdf#page=90} {Results of
  the poleval 2019 shared task 1: Recognition and normalization of temporal
  expressions}.
\newblock \emph{Proceedings ofthePolEval2019Workshop}, page~9.

\bibitem[{Lafferty et~al.(2001)Lafferty, McCallum, and
  Pereira}]{Lafferty:MaCallumg:ICML:2001:General:CRF}
John~D. Lafferty, Andrew McCallum, and Fernando C.~N. Pereira. 2001.
\newblock \href {http://dl.acm.org/citation.cfm?id=645530.655813} {Conditional
  random fields: Probabilistic models for segmenting and labeling sequence
  data}.
\newblock In \emph{Proceedings of the Eighteenth International Conference on
  Machine Learning}, ICML '01, pages 282--289, San Francisco, CA, USA. Morgan
  Kaufmann Publishers Inc.

\bibitem[{Lange et~al.(2020)Lange, Iurshina, Adel, and
  Str{\"o}tgen}]{Lange:Iurshina:RepL:2020:Adversarial/Temporal:MultilingualExtraction}
Lukas Lange, Anastasiia Iurshina, Heike Adel, and Jannik Str{\"o}tgen. 2020.
\newblock \href {https://doi.org/10.18653/v1/2020.repl4nlp-1.14} {Adversarial
  alignment of multilingual models for extracting temporal expressions from
  text}.
\newblock In \emph{Proceedings of the 5th Workshop on Representation Learning
  for NLP}, pages 103--109, Online. Association for Computational Linguistics.

\bibitem[{Laparra et~al.(2018)Laparra, Xu, and
  Bethard}]{Laparra:Xu:TACL:2018:Temporal:NeuralTempNER}
Egoitz Laparra, Dongfang Xu, and Steven Bethard. 2018.
\newblock \href {https://doi.org/10.1162/tacl_a_00025} {From characters to time
  intervals: New paradigms for evaluation and neural parsing of time
  normalizations}.
\newblock \emph{Transactions of the Association for Computational Linguistics},
  6.

\bibitem[{Lee et~al.(2014)Lee, Artzi, Dodge, and
  Zettlemoyer}]{Lee:Artzi:ACL:2014:Temporal:UWTime}
Kenton Lee, Yoav Artzi, Jesse Dodge, and Luke Zettlemoyer. 2014.
\newblock \href {https://doi.org/10.3115/v1/P14-1135} {Context-dependent
  semantic parsing for time expressions}.
\newblock In \emph{Proceedings of the 52nd Annual Meeting of the Association
  for Computational Linguistics (Volume 1: Long Papers)}.

\bibitem[{Lewis et~al.(2020)Lewis, Liu, Goyal, Ghazvininejad, Mohamed, Levy,
  Stoyanov, and Zettlemoyer}]{lewis-etal-2020-bart}
Mike Lewis, Yinhan Liu, Naman Goyal, Marjan Ghazvininejad, Abdelrahman Mohamed,
  Omer Levy, Veselin Stoyanov, and Luke Zettlemoyer. 2020.
\newblock \href {https://doi.org/10.18653/v1/2020.acl-main.703} {{BART}:
  Denoising sequence-to-sequence pre-training for natural language generation,
  translation, and comprehension}.
\newblock In \emph{Proceedings of the 58th Annual Meeting of the Association
  for Computational Linguistics}, pages 7871--7880, Online. Association for
  Computational Linguistics.

\bibitem[{Ma et~al.(2021)Ma, Zhou, Gui, Tan, Zhang, and
  Huang}]{Ma:Zhou:Arxiv:2021:NER:Prompt}
Ruotian Ma, Xin Zhou, Tao Gui, Yiding Tan, Qi~Zhang, and Xuanjing Huang. 2021.
\newblock \href {https://arxiv.org/pdf/2109.13532.pdf} {Template-free prompt
  tuning for few-shot ner}.
\newblock \emph{arXiv preprint arXiv:2109.13532}.

\bibitem[{Mazur and Dale(2010)}]{Mazur:Dale:EMNLP:2010:Temporal:WikiWars}
Pawel Mazur and Robert Dale. 2010.
\newblock \href {https://aclanthology.org/D10-1089} {{W}iki{W}ars: A new corpus
  for research on temporal expressions}.
\newblock In \emph{Proceedings of the 2010 Conference on Empirical Methods in
  Natural Language Processing}, pages 913--922, Cambridge, MA. Association for
  Computational Linguistics.

\bibitem[{Minard et~al.(2016)Minard, Speranza, Urizar, Altuna, van Erp, Schoen,
  and van Son}]{Minard:Speranza:LREC:2016:Temporal:Meantime}
Anne-Lyse Minard, Manuela Speranza, Ruben Urizar, Bego{\~n}a Altuna, Marieke
  van Erp, Anneleen Schoen, and Chantal van Son. 2016.
\newblock \href {https://aclanthology.org/L16-1699} {{MEANTIME}, the
  {N}ews{R}eader multilingual event and time corpus}.
\newblock In \emph{Proceedings of the Tenth International Conference on
  Language Resources and Evaluation ({LREC}'16)}, pages 4417--4422,
  Portoro{\v{z}}, Slovenia. European Language Resources Association (ELRA).

\bibitem[{Mirza(2016)}]{Mirza:CL:2016:Temporal:Indonesian}
Paramita Mirza. 2016.
\newblock \href {https://rd.springer.com/chapter/10.1007/978-981-10-0515-2_10}
  {Recognizing and normalizing temporal expressions in indonesian texts}.
\newblock In \emph{Computational Linguistics}, pages 135--147, Singapore.
  Springer Singapore.

\bibitem[{Orasmaa(2014)}]{Orasmaa:LREC:2014:Temporal:EstTimeML}
Siim Orasmaa. 2014.
\newblock \href
  {http://www.lrec-conf.org/proceedings/lrec2014/pdf/530_Paper.pdf} {Towards an
  integration of syntactic and temporal annotations in {E}stonian}.
\newblock In \emph{Proceedings of the Ninth International Conference on
  Language Resources and Evaluation ({LREC}'14)}, pages 1259--1266, Reykjavik,
  Iceland. European Language Resources Association (ELRA).

\bibitem[{Pustejovsky et~al.(2005)Pustejovsky, Ingria, {Saur\'i}, {Casta\~no},
  Littman, Gaizauskas, Setzer, Katz, and
  Mani}]{Pustejovsky:Ingria:OUP:2005:Temporal:TimeML}
James Pustejovsky, Robert Ingria, Roser {Saur\'i}, {Jos\'e} {Casta\~no},
  Jessica Littman, Rob Gaizauskas, Andrea Setzer, Graham Katz, and Inderjeet
  Mani. 2005.
\newblock \href
  {http://citeseerx.ist.psu.edu/viewdoc/download?doi=10.1.1.85.5610&rep=rep1&type=pdf}
  {{T}he specification language {TimeML}}.
\newblock In \emph{{T}he language of time: a reader}, pages 545--557. {O}xford
  {U}niversity {P}ress.

\bibitem[{Rahimi et~al.(2019)Rahimi, Li, and
  Cohn}]{Rahimi:Li:ACL:2019:NER:MassivelyMultilingTransfer}
Afshin Rahimi, Yuan Li, and Trevor Cohn. 2019.
\newblock \href {https://doi.org/10.18653/v1/P19-1015} {Massively multilingual
  transfer for {NER}}.
\newblock In \emph{Proceedings of the 57th Annual Meeting of the Association
  for Computational Linguistics}, pages 151--164, Florence, Italy. Association
  for Computational Linguistics.

\bibitem[{Saur{\i}(2010)}]{Sauri:Report:2010:TemporalTimeBankCA}
Roser Saur{\i}. 2010.
\newblock \href
  {https://catalog.ldc.upenn.edu/docs/LDC2012T10/annotationGuidelines_tlinks_CA_SP.pdf}
  {Annotating temporal relations in catalan and spanish timeml annotation
  guidelines}.
\newblock Technical report, Technical report, Technical Report BM 2010-04,
  Barcelona Media.

\bibitem[{Skukan et~al.(2014)Skukan, Glava{\v{s}}, and
  {\v{S}}najder}]{Skukan:Glavas:Article:2014:Temporal:CroatianRules}
Luka Skukan, Goran Glava{\v{s}}, and Jan {\v{S}}najder. 2014.
\newblock \href {http://nl.ijs.si/isjt14/proceedings/isjt2014_17.pdf}
  {Heideltime. hr: extracting and normalizing temporal expressions in
  croatian}.
\newblock In \emph{Proceedings of the 9th Slovenian Language Technologies
  Conferences (IS-LT 2014)}, pages 99--103.

\bibitem[{Str\"{o}tgen et~al.(2014)Str\"{o}tgen, Armiti, Van~Canh, Zell, and
  Gertz}]{Strotgen:Armiti:ACM:2021:Temporal:WikiWarsVi}
Jannik Str\"{o}tgen, Ayser Armiti, Tran Van~Canh, Julian Zell, and Michael
  Gertz. 2014.
\newblock \href {https://doi.org/10.1145/2540989} {Time for more languages:
  Temporal tagging of arabic, italian, spanish, and vietnamese}.
\newblock \emph{ACM Transactions on Asian Language Information Processing},
  13(1).

\bibitem[{Str{\"o}tgen and
  Gertz(2011)}]{Strotgen:Gertz:Paper:2011:Temporal:WikiWarsDE}
Jannik Str{\"o}tgen and Michael Gertz. 2011.
\newblock \href
  {http://citeseerx.ist.psu.edu/viewdoc/download?doi=10.1.1.384.6136&rep=rep1&type=pdf#page=135}
  {Wikiwarsde: A german corpus of narratives annotated with temporal
  expressions}.
\newblock In \emph{Proceedings of the conference of the German society for
  computational linguistics and language technology (GSCL 2011)}, pages
  129--134. Citeseer.

\bibitem[{Str\"{o}tgen and
  Gertz(2013)}]{Strotgen:Gertz:LRE:2013:Temporal:MultilingDomainTagging}
Jannik Str\"{o}tgen and Michael Gertz. 2013.
\newblock \href {https://doi.org/10.1007/s10579-012-9179-y} {Multilingual and
  cross-domain temporal tagging}.
\newblock \emph{Language Resources and Evaluation}, 47(2).

\bibitem[{Str{\"o}tgen and
  Gertz(2015)}]{Streotgen:Gertz:EMNLP:2015:Temporal:BaselineTemporalTagger}
Jannik Str{\"o}tgen and Michael Gertz. 2015.
\newblock \href {https://doi.org/10.18653/v1/D15-1063} {A baseline temporal
  tagger for all languages}.
\newblock In \emph{Proceedings of the 2015 Conference on Empirical Methods in
  Natural Language Processing}.

\bibitem[{Str{\"o}tgen and
  Gertz(2016)}]{Strotgen:Gertz:Book:2016:Temporal:Tagging}
Jannik Str{\"o}tgen and Michael Gertz. 2016.
\newblock \href {https://doi.org/10.2200/S00721ED1V01Y201606HLT036}
  {\emph{Domain-sensitive temporal tagging}}, volume~9.
\newblock Morgan \& Claypool Publishers.

\bibitem[{Str{\"o}tgen et~al.(2018)Str{\"o}tgen, Minard, Lange, Speranza, and
  Magnini}]{Strotgen:Minard:LREC:2018:Temporal:Krauts}
Jannik Str{\"o}tgen, Anne-Lyse Minard, Lukas Lange, Manuela Speranza, and
  Bernardo Magnini. 2018.
\newblock \href {https://aclanthology.org/L18-1085} {{KRAUTS}: A {G}erman
  temporally annotated news corpus}.
\newblock In \emph{Proceedings of the Eleventh International Conference on
  Language Resources and Evaluation ({LREC} 2018)}, Miyazaki, Japan. European
  Language Resources Association (ELRA).

\bibitem[{Sun et~al.(2021)Sun, Liu, Qiu, and
  Huang}]{Sun:Liu:Arxiv:2021:General:Paradigmshift}
Tianxiang Sun, Xiangyang Liu, Xipeng Qiu, and Xuanjing Huang. 2021.
\newblock \href {https://arxiv.org/pdf/2109.12575.pdf} {Paradigm shift in
  natural language processing}.
\newblock \emph{arXiv preprint arXiv:2109.12575}.

\bibitem[{UzZaman et~al.(2013)UzZaman, Llorens, Derczynski, Allen, Verhagen,
  and Pustejovsky}]{Uzzaman:Llorens:SEM:2013:Temporal:TempEval3}
Naushad UzZaman, Hector Llorens, Leon Derczynski, James Allen, Marc Verhagen,
  and James Pustejovsky. 2013.
\newblock \href {https://aclanthology.org/S13-2001} {{S}em{E}val-2013 task 1:
  {T}emp{E}val-3: Evaluating time expressions, events, and temporal relations}.
\newblock In \emph{Second Joint Conference on Lexical and Computational
  Semantics (*{SEM}), Volume 2: Proceedings of the Seventh International
  Workshop on Semantic Evaluation ({S}em{E}val 2013)}, pages 1--9, Atlanta,
  Georgia, USA. Association for Computational Linguistics.

\end{thebibliography}
\bibliographystyle{acl_natbib}


\appendix

\section{Slot Tokenization of CIRs }\label{sec:appendix:values}
In this section, we describe our slot-based tokenization of the context-independent representation (CIR) of values as introduced in Section~\ref{sec:background} and Section~\ref{sec:approach} of the main paper. 

\lstdefinestyle{custom}{
    language=sh,
    basicstyle=\ttfamily,
    keywordstyle=\color{Black}\bf,
    morekeywords={SB, SB, SD1, SD2, SD3, SD4, ST1, ST2, ST3, SA1, SA2, SA3, SA3} 
    literate={\ \ }{{\ }}1,
}
\lstset{style=custom}

\subsection{Overview of Slots}
We use the following 11 slots to represent CIRs values.\footnote{Note that our CIRs describe a superset of TimeML.} These slots are then used for masking during training and inference with our normalization model (which basically is a masked language model).

\paragraph{SB:} This slot can contain BC information of years (e.g., as in \texttt{BC4000} for the year 4000 BC) or the duration markers P and PT. Moreover, mathematical operations like \texttt{PLUS} are covered as used in relative expression involving offset computations (e.g., \texttt{this-day-plus-2} for the day after tomorrow) and holiday names (\texttt{EasterSunday}).
    
\paragraph{SD1, SD2:} These slots are used to represent 4-digit year numbers (\textbf{SD1} = \texttt{20} and \textbf{SD2} = \texttt{22} for the year \texttt{2022}) by splitting the 4-digit number into two 2-digit numbers. 
This helps to generalize to unseen years as fewer parameters have to be learned. In addition, we use \textbf{SD1} to mark reference expressions like \texttt{PAST\_REF}. 
For underspecifed expressions like \texttt{UNDEF-this-day}, \texttt{this} is stored in \textbf{SD1} and \texttt{day} in \textbf{SD2}. Moreover, \textbf{SD1} and \textbf{SD2} are used to store numbers of \texttt{DURATION} expressions. 
    
\paragraph{SD3, SD4:} Analogously to \textbf{SD1} and \textbf{SD2} that are used to store year information, \textbf{SD3} is used for months and \textbf{SD4} for days. 
    
\paragraph{ST1, ST2, ST3:} Temporal information from expressions of type \texttt{TIME} that are smaller than day granularity are stored in the \textbf{ST} slots. For example, the hour information of \texttt{24:00} and the daytime information, such as \texttt{EV} is stored in \textbf{ST1}. Information on minutes and seconds is stored in \textbf{ST2} and \textbf{ST3}, respectively. Moreover, these slots are used to cover additional units in durations, such as in \texttt{P1D2H} (1 day and 2 hours). 
    
\paragraph{SA1, SA2, SA3:} Finally, some CIRs include function calls which can be augmented with arguments that we store in the \textbf{SA} slots. For example, the argument \texttt{2} of \texttt{this-day-plus-2} is stored in \textbf{SA1}. Other function calls are used to compute days with respect to holidays like \texttt{EaserSunday} or specific weekdays. 

Note that slots can be optional depending on the temporal expression. For example, the value \texttt{2022} representing the year 2022 would only require \textbf{SD1} and \textbf{SD2}. All other slots are set to a padding value \texttt{[PAD]} then which allows a fixed-sized representations of CIRs that can be predicted with our masked language model. 

\newcommand{\teArrow}{$\rightarrow$~}

\paragraph{Examples. }
The following examples show temporal expressions, their corresponding CIRs and the tokenization into our slots. 
Note that there is no need to capture terms like \texttt{UNDEF} in our slots as the presence of words like \texttt{this}, \texttt{next} or \texttt{last} in a CIR implies the existence of \texttt{UNDEF} 
in the CIR. This information can be reconstructed when obtaining a CIR from our slots. This also includes \texttt{--} to separate numbers as in \texttt{YYYY-MM-DD} values, \texttt{REF} in reference expressions and \texttt{T} for time information. 
We use the following format to give examples for our CIR conversion:  \textit{Text} \teArrow \textit{CIR} \teArrow \textit{Slot Sequence} 

\begin{itemize}
    \item \emph{Now} ... 
    \\ \teArrow \texttt{PRESENT\_REF} 
    \\ \teArrow \textbf{SD1}=\texttt{PRESENT}
    
    \item ... for \emph{1000 days} ... 
    \\ \teArrow \texttt{P1000D} 
    \\ \teArrow \textbf{SB}=\texttt{P}, \textbf{SD1}=\texttt{10}, \textbf{SD2}=\texttt{00}, \textbf{SD4}=\texttt{D}
    
    \item ... for \emph{one and a half day} ... 
    \\ \teArrow \texttt{P1D12H}
    \\ \teArrow \textbf{SB}=\texttt{P}, \textbf{SD1}=\texttt{1}, \textbf{SD4}=\texttt{D}, \textbf{ST1}=\texttt{12}, \textbf{ST2}=\texttt{H}
    
    \item ... in \emph{1000 BC} ... 
    \\ \teArrow \texttt{BC1000}
    \\ \teArrow \textbf{SB}=\texttt{BC}, \textbf{SD1}=\texttt{10}, \textbf{SD2}=\texttt{00}
    
    \item ... on the \emph{morning of March 15, 2022} ...
    \\ \teArrow \texttt{2022-03-15TMO}
    \\ \teArrow \textbf{SD1}=\texttt{20}, \textbf{SD2}=\texttt{22}, \textbf{SD3}=\texttt{03}, \textbf{SD4}=\texttt{15}, \textbf{ST1}=\texttt{MO}
    
    \item On \emph{March 15}, ... 
    \\ \teArrow \texttt{UNDEF-year-03-15}
    \\ \teArrow \textbf{SD1}=\texttt{year}, \textbf{SD3}=\texttt{03}, \textbf{SD4}=\texttt{15}
    
    \item ... \emph{the day after tomorrow} ...
    \\ \teArrow \texttt{UNDEF-this-day-PLUS-2}
    \\ \teArrow \textbf{SB}=\texttt{PLUS}, \textbf{SD1}=\texttt{this}, \textbf{SD2}=\texttt{day}, \textbf{SA1}=\texttt{2}
    
    \item ... at \emph{Pentecost}\footnote{In christian communities, the holiday of Pentecost is celebrated 49 days after Easer Sunday.} ... 
    \\ \teArrow \texttt{UNDEF-year-00-00 funcDate}... ...\texttt{Calc(EasterSunday(YEAR, 49))}
    \\ \teArrow \textbf{SB}=\texttt{EasterSunday}, \textbf{SD1}=\texttt{year}, \textbf{SD2}=\texttt{00},  \textbf{SA1}=\texttt{49}
\end{itemize}

\subsection{Regular expressions}
In the following, we will describe the six regular expressions used to split CIR values from \heideltime outputs into our slots for the weakly-supervised training data. 
    
\paragraph{Notation.}
For readability, we define the following groups to capture temporal units and other fixed names. Note that these are used across languages. For example, the German expression \textit{Montag} would still be represented with \texttt{monday}. 

\begin{lstlisting}
UNITS = (H|D|DE|DT|M|C|Y|
    C|CE|W|WE|Qu|Q|S)
UNITS_F = (day|month|year|
    decade|century|week|
    weekend|quarter|
    hour|minute|second)
DAYTIME = (NI|AF|MO|EV|MD|MI)
SPECIAL = (SP|SU|FA|AU|WI|
    H1|H2|Q1|Q2|Q3|Q4|H|Q)
NAMES = (monday|tuesday|
    wednesday|thursday|
    friday|saturday|sunday|
    january|february|march|
    april|may|june|july|
    august|september|
    october|november|december)
\end{lstlisting}

In the following, $DX(n)$ marks the $n$-th group captured by the regular expression $DX$.

\paragraph{$D1$: References. }
The first regular expression $D1$ is used to capture simple reference expressions that refer to uncertain points in time. 
    
\begin{lstlisting}
D1 = (PRESENT|PAST|FUTURE)_REF
Slots: SD1=D1(1) 
\end{lstlisting}

\paragraph{$D2$: Explicit Dates. }
The second regular expression $D2$ detects explicit values that do not need further normalization, such as days in the \texttt{YYYY-DD-MM} format, e.g., \texttt{20222-03-15}. 
    
\begin{lstlisting}
D2 =(BC)?(\d\d?|XX)?(\d\d|XX)?
  (?:-(W)?(\d\d?|XX|SPECIAL))?
  (?:-(\d\d?|XX|WE))?\)?
  (?:T(\d\d|X|DAYTIME|XX)?
  (?::(\d\d))?
  (?:(?::|-)(\d\d))?)?
Slots: SB=D2(1), SD1=D2(2), 
  SD2=D2(3), SD3=D2(5), SD4=D2(6), 
  ST1=D2(7), ST2=D2(8), 
  ST3=D2(9)|D2(4)
\end{lstlisting}

\paragraph{$P1$: Durations. }
The third regular expression $P1$ detects expressions of \texttt{type} \textsc{Duration}, e.g., \texttt{P1D2H}. 
These are defined as \texttt{P<number><unit>} for units of at least day granularity and \texttt{PT<number><unit>} for smaller granularities. We capture up to two different units \texttt{P1D2H} (1 day and 2 hours) but ignore further units that are theoretically defined in the TimeML specifications but do not occur often in practice (in our datasets those did not occur at all). 
    
\begin{lstlisting}
P1 = (P|PT)(\d\d?|X|XX)
  (\d\d|\.)?(\d\d?)?)?(UNITS)?
  (\d\d?)?(UNITS)?
Slots: SB=P1(1), SD1=P1(2), 
  SD2=P1(3), SD3=P1(4), SD4=P1(6), 
  ST1=P1(5), ST2=P1(7)
\end{lstlisting}

\paragraph{$D3$: Relative Dates. }
While the previous regular expressions $D1$, $D2$ and $P1$ follow the TimeML specifications and capture fully normalized expressions, i.e., anchored values, the following regular expressions capture CIRs as used internally by \heideltime. They represent relative expressions that need to be anchored. 

$D3$ detects relative expressions with respect to a certain point in time, such as \texttt{this-day-plus-2} (the day after tomorrow). 

\begin{lstlisting}
D3 = UNDEF-(this|next|last|REF|
      REFUNIT|REFDATE)?-?
  (UNITS_F|SPECIAL)?-??
  (NAMES|SPECIAL)|XX|\d\d?)?
  (?:-?(\d\d?|XX))?
  (?:-(PLUS|MINUS|LESS)-(\d\d?)-?
  (\d\d?)?-?(\d\d?)?)?\)?
  (?:T(\d\d?|X|DAYTIME|XX)?
  (?::(\d\d?|XX))?(?:(?::|-)
      (\d\d|XX))?)
Slots: SB=D3(5), SD1=D3(1), 
  SD2=D3(2), SD3=D3(3), SD4=D3(4), 
  ST1=D3(9),ST2=D3(10),ST3=D3(11),
  SA1=D3(6), SA2=D3(7), SA3=D3(8)
\end{lstlisting}

\paragraph{$D4$: Relative Dates (coarse). }
$D4$ captures underspecified expressions like \textit{May} that is missing year information and would be represented with the CIR \texttt{UNDEF-year-05}. 

\begin{lstlisting}
D4 = UNDEF-(year|decade|century?)
  -?(\d\d?|X)?-?(\d\d?|X)?-?
  (\d\d?|X|SPECIAL)?\)?
  (?:T(\d\d?|X|DAYTIME)?
  (?::(\d\d?|XX))?
  (?:(?::|-)(\d\d|XX))?)?
Slots: SD1=D4(1), SD2=D4(2),
  SD3=D4(3), SD4=D4(4), 
  ST1=D4(5), ST2=D4(6), ST3=D4(7)
\end{lstlisting}

\paragraph{$D5$: Holidays and functions. }
Finally, $D5$ covers special functions used by \heideltime. 
These functions are used to compute days with respect to weekdays and moveable feasts like \texttt{EasterSunday} that refer to different days depending on the year. For example, the earliest possible date of Easter Sunday is March 22 and the latest is April 25 in the Gregorian calendar.\footnote{\url{https://en.wikipedia.org/wiki/List_of_dates_for_Easter}} The concrete date is then computed by an external function given a year.\footnote{\url{https://www.linuxtopia.org/online_books/programming_books/python_programming/python_ch38.html}}

\begin{lstlisting}
D5 = (UNDEF-year|UNDEF-this-year|
        UNDEF-century\d\d|\d\d\d\d)-
    (\d\d)-00 funcDateCalc\((
        WeekdayRelativeTo|
        EasterSundayOrthodox|
        EasterSunday|
        ShroveTideOrthodox)
    \(YEAR(?:(?:-(\d\d)))?
    (?:-(\d\d))
    (?:,\s?(-?\d\d?))?
    (?:,\s?(-?\d\d?))?
    (?:, (true|false))?\)\)
Slots: SB=D5(3), SD1=D4(1), 
    SD2=D4(2), SD3=D4(5), 
    SD4=D4(7), ST1=D4(3), 
    SA1=D5(6), SA2=D5(7), SA3=D5(8)
\end{lstlisting}

\section{Data Statistics}

\subsection{Weakly-Supervised Data}\label{app:sec:weak}
As detailed in Section~\ref{sec:experiments}, we create weakly-supervised data to train our normalization model, as the gold standard is too small and is not annotated with CIRs which are required by our method.
For all languages, we take the data from GlobalVoices\footnote{\url{https://globalvoices.org/}} (news-style documents) and Wikipedia\footnote{\url{https://en.wikipedia.org/wiki/List_of_Wikipedias}} (narrative-style documents), use \spacy for tokenization and our \heideltime version that outputs CIRs for the annotation with temporal expressions. 
The sizes of our weakly-supervised data for each language are given in Table~\ref{tab:data_weak}.

\subsection{Gold-Standard Data}\label{app:sec:gold}
Detailed information on the datasets used in this paper (their languages, domains, sizes and references) are provided in Table~\ref{tab:datasets}. 
Note that all corpora come from the news domain except the WikiWars corpora that are based on Wikipedia articles.

\subsection{Distribution of Explicit and Relative Values}\label{sec:app:dist}
The distribution of explicit and relative values has a large impact on the normalization performance of different models, as shown in our ablation study in Section~\ref{sec:ablation}. 
Exemplarily, we analyze their distribution in the German and English datasets for which we have data from two domains: News and Wikipedia. The results are given in Table~\ref{tab:value_distribution}. We see, that the Wikipedia corpora contain a much larger percentage of relative values as these articles often follow a narrative structure (cf., \citep{Strotgen:Gertz:Book:2016:Temporal:Tagging}.

\begin{table}
\centering
\begin{tabular}{l|cc} \toprule
 & De & En \\ \midrule
News & \textbf{67.1} / 32.9 & \textbf{52.3} / 47.7 \\
Wiki & 47.6 / \textbf{52.4} & 44.2 / \textbf{55.8} \\ \bottomrule
\end{tabular}
\caption{Distribution of explicit / relative values according to \heideltime by domains (in \%). }
\label{tab:value_distribution}
\end{table}

\section{A Note on Adopting \heideltime}
In our experiments, we used a modified version of \heideltime. 
First, we implemented a new UIMA collection reader based on \spacy as an alternative to the TreeTagger that has a restrictive license. This results in a slightly different sentence segmentation and tokenization, and, thus, minor differences in performance. For example, the original \heideltime achieves 63.47 \fscore on the Portuguese test data, while our \spacy version achieves 63.24 \fscore as one additional false positive expression was annotated due to different sentence boundaries. 
Second, we adapted HeidelTime to output its internal CIRs for the TimeML values, such that we can create our weakly-supervised training data.

The rather low performance of our models and \heideltime for the high-resource languages Estonian (et) and Dutch (nl) can be explained by poor data quality. 
An inter-annotator agreement of 44 \fscore was reported for the Estonian corpus \cite{Orasmaa:LREC:2014:Temporal:EstTimeML}, which is close to our results. The Dutch data was translated from English and automatically annotated via cross-lingual projections \cite{Minard:Speranza:LREC:2016:Temporal:Meantime}, which may reduce the annotation quality. Note, that only the first five sentences for each document were annotated in the Meantime corpora (it and nl). We restricted our evaluation to these annotated parts accordingly.

\newcommand{\langAF}{af}
\newcommand{\langAM}{am}
\newcommand{\langAR}{ar}
\newcommand{\langAS}{as}
\newcommand{\langBG}{bg}
\newcommand{\langBN}{bn}
\newcommand{\langBR}{br}
\newcommand{\langCA}{ca}
\newcommand{\langCS}{cs}
\newcommand{\langCY}{cy}
\newcommand{\langDA}{da}
\newcommand{\langDE}{de}
\newcommand{\langEL}{el}
\newcommand{\langEN}{en}
\newcommand{\langEO}{eo}
\newcommand{\langES}{es}
\newcommand{\langET}{et}
\newcommand{\langEU}{eu}
\newcommand{\langFA}{fa}
\newcommand{\langFI}{fi}
\newcommand{\langFR}{fr}
\newcommand{\langFY}{fy}
\newcommand{\langGA}{ga}
\newcommand{\langGD}{gd}
\newcommand{\langGL}{gl}
\newcommand{\langGU}{gu}
\newcommand{\langHA}{ha}
\newcommand{\langHE}{he}
\newcommand{\langHI}{hi}
\newcommand{\langHR}{hr}
\newcommand{\langHU}{hu}
\newcommand{\langHY}{hy}
\newcommand{\langID}{id}
\newcommand{\langIS}{is}
\newcommand{\langIT}{it}
\newcommand{\langJA}{ja}
\newcommand{\langJV}{jv}
\newcommand{\langKA}{ka}
\newcommand{\langKK}{kk}
\newcommand{\langKM}{km}
\newcommand{\langKN}{kn}
\newcommand{\langKO}{ko}
\newcommand{\langKU}{ku}
\newcommand{\langKY}{ky}
\newcommand{\langLA}{la}
\newcommand{\langLO}{lo}
\newcommand{\langLT}{lt}
\newcommand{\langLV}{lv}
\newcommand{\langMG}{mg}
\newcommand{\langMK}{mk}
\newcommand{\langML}{ml}
\newcommand{\langMN}{mn}
\newcommand{\langMR}{mr}
\newcommand{\langMS}{ms}
\newcommand{\langMY}{my}
\newcommand{\langNE}{ne}
\newcommand{\langNL}{nl}
\newcommand{\langOA}{oa}
\newcommand{\langOM}{om}
\newcommand{\langOR}{or}
\newcommand{\langPL}{pl}
\newcommand{\langPS}{ps}
\newcommand{\langPT}{pt}
\newcommand{\langRO}{ro}
\newcommand{\langRU}{ru}
\newcommand{\langSA}{sa}
\newcommand{\langSD}{sd}
\newcommand{\langSI}{si}
\newcommand{\langSK}{sk}
\newcommand{\langSL}{sl}
\newcommand{\langSO}{so}
\newcommand{\langSQ}{sq}
\newcommand{\langSR}{sr}
\newcommand{\langSU}{su}
\newcommand{\langSV}{sv}
\newcommand{\langSW}{sw}
\newcommand{\langTA}{ta}
\newcommand{\langTH}{th}
\newcommand{\langTR}{tr}
\newcommand{\langUG}{ug}
\newcommand{\langUK}{uk}
\newcommand{\langUR}{ur}
\newcommand{\langUZ}{uz}
\newcommand{\langVI}{vi}
\newcommand{\langXH}{xh}
\newcommand{\langYI}{yi}
\newcommand{\langYO}{yo}
\newcommand{\langZH}{zh}

\begin{table*}
\begin{subtable}[t]{0.30\linewidth}
\centering
\begin{tabular}{llr} \toprule
Rank & Lang & \#Ann. \\ \midrule
1 & \langDE & 870897 \\
2 & \langEN & 542087 \\
3 & \langFR & 284871 \\
4 & \langAR & 280446 \\
5 & \langES & 250871 \\
6 & \langPT & 215209 \\
7 & \langIT & 199236 \\
8 & \langNL & 194944 \\
9 & \langRU & 122884 \\
10 & \langZH & 105421 \\
11 & \langHR & 50233 \\
12 & \langRO & 33545 \\
13 & \langVI & 22048 \\
14 & \langAF & 21081 \\
15 & \langMK & 19539 \\
16 & \langTR & 19532 \\
17 & \langGL & 17416 \\
18 & \langCA & 16747 \\
19 & \langBN & 16284 \\
20 & \langCY & 14738 \\
21 & \langBG & 14550 \\
22 & \langET & 13948 \\
23 & \langSV & 13705 \\
24 & \langID & 13031 \\
25 & \langDA & 12919 \\
26 & \langFY & 12852 \\
27 & \langPL & 11283 \\
28 & \langFA & 11041 \\
29 & \langEU & 10992 \\\bottomrule
\end{tabular}
\end{subtable}
\hfill
\begin{subtable}[t]{0.30\linewidth}
\centering
\begin{tabular}{llr} \toprule
Rank & Lang & \#Ann. \\ \midrule
30 & \langNE & 10750 \\ 
31 & \langMS & 10017 \\
32 & \langMG & 9271 \\
33 & \langKK & 8080 \\
34 & \langHI & 7762 \\
35 & \langEO & 7353 \\
36 & \langUR & 6228 \\
37 & \langHU & 5871 \\
38 & \langSQ & 5760 \\
39 & \langSK & 5172 \\
40 & \langSR & 4276 \\
41 & \langKA & 4247 \\
42 & \langEL & 4217 \\
43 & \langHE & 4057 \\
44 & \langSW & 3979 \\
45 & \langJA & 3696 \\
46 & \langBR & 3582 \\
47 & \langUZ & 3361 \\
48 & \langTH & 3162 \\
49 & \langCS & 3096 \\
50 & \langGA & 2799 \\
51 & \langMN & 2778 \\
52 & \langGD & 2772 \\
53 & \langLT & 2734 \\
54 & \langMR & 2623 \\
55 & \langLA & 1876 \\
56 & \langUK & 1673 \\
57 & \langHY & 1642 \\ 
58 & \langTA & 1556 \\ \bottomrule
\end{tabular}
\end{subtable}
\hfill
\begin{subtable}[t]{0.30\linewidth}
\centering
\begin{tabular}{llr} \toprule
Rank & Lang & \#Ann. \\ \midrule
59 & \langMY & 1103 \\
60 & \langML & 1079 \\
61 & \langKN & 1029 \\
62 & \langFI & 1017 \\
63 & \langOA & 979 \\
64 & \langJV & 968 \\
65 & \langKY & 926 \\
66 & \langIS & 804 \\
67 & \langAM & 776 \\
68 & \langKU & 557 \\
69 & \langSO & 506 \\
70 & \langYI & 485 \\
71 & \langKO & 483 \\
72 & \langSI & 442 \\
73 & \langPS & 403 \\
74 & \langLO & 354 \\
75 & \langKM & 350 \\
76 & \langSU & 335 \\
77 & \langLV & 323 \\
78 & \langAS & 299 \\
79 & \langUG & 283 \\
80 & \langSD & 278 \\
81 & \langGU & 258 \\
82 & \langHA & 205 \\
83 & \langSL & 125 \\
84 & \langYO & 102 \\
85 & \langSA & 24 \\
86 & \langOR & 19 \\
87 & \langXH & 3 \\  \bottomrule
\end{tabular}
\end{subtable}
\caption{Languages and the sizes of our weakly-supervised data. }
\label{tab:data_weak}
\end{table*}

\begin{table*}
\centering
\begin{tabular}{llll}
\toprule
Corpus & Language & \begin{tabular}[c]{@{}l@{}}\#Annotations\\ (train / test)\end{tabular} & Reference \\ \midrule
\multicolumn{4}{l}{\quad \textit{Corpora only used for evaluation}} \\
KRAUTS-DieZeit         & German (de)    & \_ / 493    & \cite{Strotgen:Minard:LREC:2018:Temporal:Krauts} \\
TempEval-3 (platinum)  & English (en)   & \_ / 137    & \cite{Uzzaman:Llorens:SEM:2013:Temporal:TempEval3} \\
KOMPAS (test)          & Indonesian (id) & \_ / 192    & \cite{Mirza:CL:2016:Temporal:Indonesian} \\
TimeBankCA             & Catalan (ca)   & \_ / 1383   & \cite{Sauri:Report:2010:TemporalTimeBankCA} \\
EstTimeML              & Estonian (et)  & \_ / 622    & \cite{Orasmaa:LREC:2014:Temporal:EstTimeML} \\
EusTimeML              & Basque (eu)    & \_ / 112    & \cite{Altuna:Aranzabe:Article:2020:TemporalEusTimeML} \\
Fr TimeBank            & French (fr)     & \_ / 423    & \cite{Bittar:Amsili:ACL:2011:Temporal:FrenchTimeBank} \\
Ro TimeBank             & Romanian (ro)  & \_ / 151    & \cite{Forascu:Tufis:LREC:2012:Temporal:RomanianTimeBank} \\
PT-TimeBank (test)     & Portuguese (pt) & \_ / 151    & \cite{Costa:Branco:LREC:2012:Temporal:TimeBankPT} \\ 
WikiWars-EL (test)     & Greek (el)     & \_ / 414 &
\cite{Kapernaros:Thesis:2020:Temporal:WikiWarsEL} \\ \midrule
\multicolumn{4}{l}{\quad \textit{Corpora split into train and test sets}} \\
Meantime (IT)          & Italian (it)    & 229 / 244   & \cite{Minard:Speranza:LREC:2016:Temporal:Meantime} \\
Meantime (NL)          & Dutch (nl)     & 221 / 259   & \cite{Minard:Speranza:LREC:2016:Temporal:Meantime} \\
TempEval-3 (ES)        & Spanish (es)    & 730 / 551   & \cite{Uzzaman:Llorens:SEM:2013:Temporal:TempEval3} \\
PolEval-2019           & Polish (pl)     & 633 / 6011  & \cite{Kocon:Oleksy:Poleval:2019:Temporal:Overview} \\
WikiWars               & English (en)   & 1378 / 1251 & \cite{Mazur:Dale:EMNLP:2010:Temporal:WikiWars} \\
WikiWars-DE            & German (de)     & 1510 / 684  & \cite{Strotgen:Gertz:Paper:2011:Temporal:WikiWarsDE} \\
WikiWars-HR            & Croatian (hr)   & 724 / 677   & \cite{Skukan:Glavas:Article:2014:Temporal:CroatianRules} \\
WikiWars-UA            & Ukrainian (ua)  & 454 / 2237  & \cite{Grabar:Hamon:Article:2019:Temporal:WikiwarsUA} \\
WikiWars-VI            & Vietnamese (vi) & 118 / 101   & \cite{Strotgen:Armiti:ACM:2021:Temporal:WikiWarsVi} \\ \midrule
\multicolumn{4}{l}{\quad \textit{Corpora only used for training}} \\
KRAUTS-Dolomiten       & German (de)     & 388 / \_    & \cite{Strotgen:Minard:LREC:2018:Temporal:Krauts} \\
Meantime (EN)          & English (en)   & 472 / \_    & \cite{Minard:Speranza:LREC:2016:Temporal:Meantime} \\
TempEval-3 (train, \langEN) & English (en)    & 1240 / \_   & \cite{Uzzaman:Llorens:SEM:2013:Temporal:TempEval3} \\
PT-TimeBank (train)    & Portuguese (pt) & 1127 / \_  & \cite{Costa:Branco:LREC:2012:Temporal:TimeBankPT} \\ 
WikiWars-EL (train)    & Greek (el)      & 1496 / \_  & \cite{Kapernaros:Thesis:2020:Temporal:WikiWarsEL} \\ \bottomrule
\end{tabular}
\caption{Overview of datasets and details on their usage as training (extraction-only) or evaluation data. }
\label{tab:datasets}
\end{table*}

\end{document}